\documentclass[10pt,twocolumn,letterpaper]{article}

\usepackage[pagenumbers]{wacv} 

\definecolor{wacvblue}{rgb}{0.21,0.49,0.74}
\usepackage[pagebackref,breaklinks,colorlinks,allcolors=wacvblue]{hyperref}
\usepackage{multirow}

\def\wacvPaperID{1141} 
\def\confName{WACV}
\def\confYear{2026}

\title{PALMS+: Modular Image-Based Floor Plan Localization Leveraging Depth Foundation Model}

\author{
Yunqian Cheng \quad Benjamin Princen \quad Roberto Manduchi \\ \\
University of California, Santa Cruz \\
Santa Cruz, United States \\
{\tt\small \{ychen827, bprincen, manduchi\}@ucsc.edu}
}

\begin{document}
\maketitle
\begin{abstract}

Indoor localization in GPS-denied environments is crucial for applications like emergency response and assistive navigation. Vision-based methods such as PALMS enable infrastructure-free localization using only a floor plan and a stationary scan, but are limited by the short range of smartphone LiDAR and ambiguity in indoor layouts. We propose PALMS+, a modular, image-based system that addresses these challenges by reconstructing scale-aligned 3D point clouds from posed RGB images using a foundation monocular depth estimation model (Depth Pro), followed by geometric layout matching via convolution with the floor plan. PALMS+ outputs a posterior over the location and orientation, usable for direct or sequential localization. Evaluated on the Structured3D and a custom campus dataset consisting of 80 observations across four large campus buildings, PALMS+ outperforms PALMS and F$^3$Loc in stationary localization accuracy—without requiring any training. Furthermore, when integrated with a particle filter for sequential localization on 33 real-world trajectories, PALMS+ achieved lower localization errors compared to other methods, demonstrating robustness for camera-free tracking and its potential for infrastructure-free applications. Code and data are available at \url{https://github.com/Head-inthe-Cloud/PALMS-Plane-based-Accessible-Indoor-Localization-Using-Mobile-Smartphones}.
\end{abstract}
    
\section{Introduction}
\label{sec:intro}

\begin{figure*}
  \centering
  \includegraphics[width=1\linewidth]{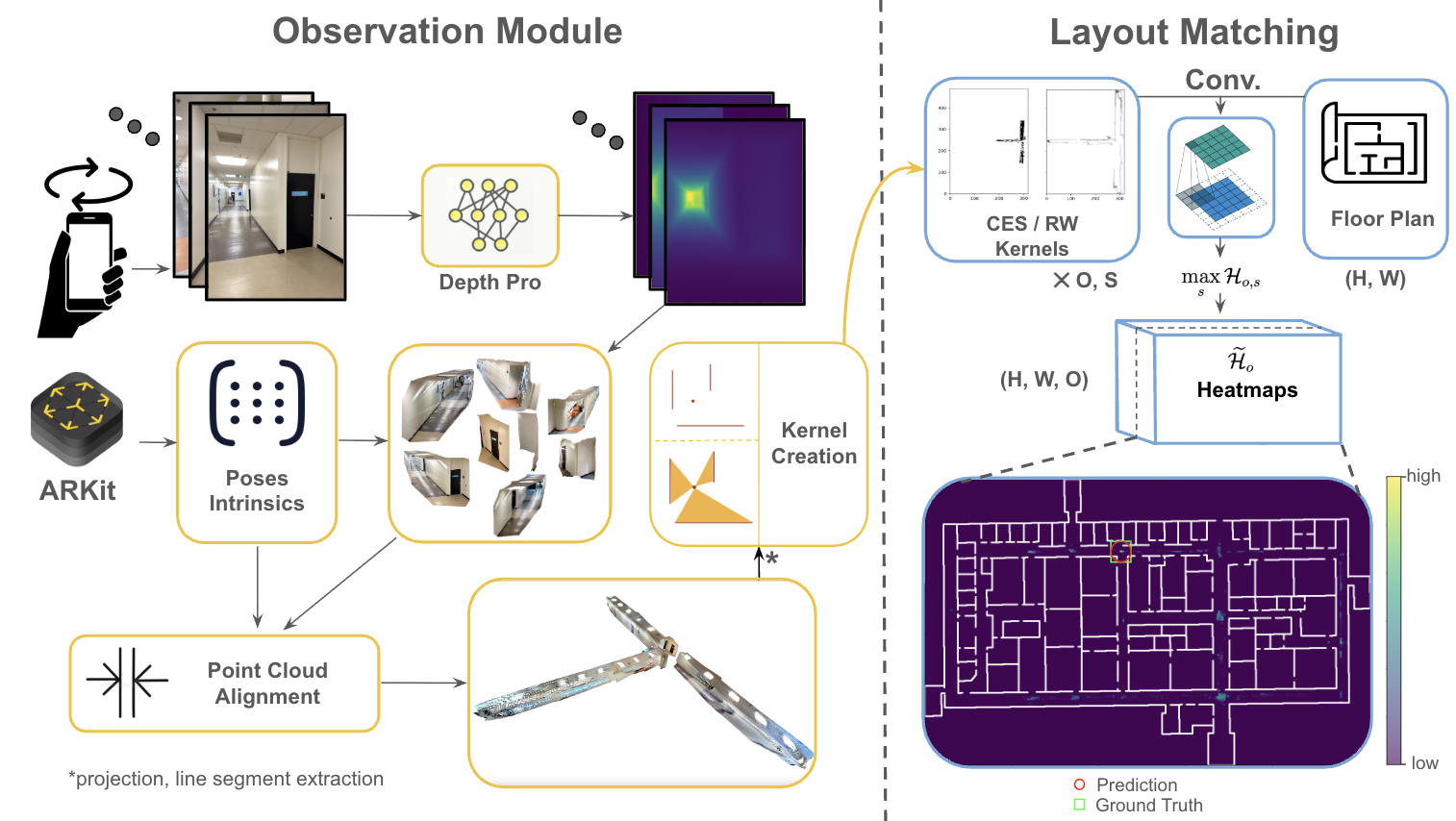}
  \caption{The PALMS+ image-based floor plan localization method.}
  \label{fig:palms+}
\end{figure*}

When visiting a large indoor space for the first time, a floor plan with a ``You are here" marker makes localization easy. Without it, one must interpret surrounding geometry—hallways, doorways, wall layouts—and match it to the map (we refer to this as {\em layout matching}), which can be ambiguous in repetitive environments.

Indoor localization addresses this challenge, enabling users to identify their location within a map in an unfamiliar, GPS-denied environment. This capability has critical applications ranging from emergency response to improving the independence of blind and visually impaired (BVI) individuals. Among various approaches,  vision-based methods \cite{howard2021lalaloc, howard2022lalaloc++, min2022laser, chen2024f3loc, cheng2024palms, li2025flona, Karkus2018ParticleFN} offer an attractive, low-cost solution by avoiding the need for additional infrastructure or prior in-building data collection, relying instead on widely available floor plans.

Prior work like PALMS~\cite{cheng2024palms} has demonstrated this idea by enabling users to take a stationary LiDAR scan of their surroundings with a smartphone, finding one or more possible locations via layout matching, then resolving any ambiguities using a particle filtering algorithm~\cite{Dellaert1999MonteCL}. However, PALMS is limited by the short range of smartphone LiDAR (5m on iPhone 14 Pro) and the inherent geometric ambiguity of indoor spaces.

To address these limitations, we present \textbf{PALMS+}, an image-based indoor localization system designed to improve localization accuracy and reduce reliance on subsequent motion. PALMS+ is composed of two key modules: 1) an \textbf{observation module} that uses state-of-the-art monocular depth estimation  (Depth Pro~\cite{Bochkovskiy2024DepthPS}) to reconstruct a scale-aligned 3D point cloud from one or more images (with estimated camera poses) taken with a standard smartphone camera,  and 2) a \textbf{layout matching module} adapted from PALMS, which performs geometric-feature-based matching by convolving the projected point cloud with the floor plan to estimate a probability distribution over possible poses. The full system is illustrated in \cref{fig:palms+}.

PALMS+ is designed to be modular and broadly applicable. Unlike PALMS, it does not require on-device LiDAR, making it compatible with virtually any modern smartphone or wearable device with a camera. Since PALMS+ doesn't require any model training, its observation and layout matching modules can be interchanged or upgraded with alternative depth estimation, camera-pose-tracking, or observation-to-floor-plan-matching methods. Moreover, PALMS+ requires only a floor plan of the building, something that in most cases is publicly available.

The output of the two modules in PALMS+ is a three-dimensional (x-y user location and orientation) posterior density, sometimes referred to as a {\em heatmap}. One could simply select the argmax as the predicted pose, as done in ~\cite{chen2024f3loc, howard2022lalaloc++}. However, inherent geometric ambiguities often produce multimodal heatmaps, where poses with geometry resembling the true pose—sometimes far away from it—also receives high posterior density. To mitigate the risk of incorrect mode selection, sequential localization can be employed. In our study, we emphasize PALMS+'s ability to localize from a single stationary observation, while also demonstrating that its heatmap output integrates naturally with a particle filter for sequential localization, yielding improved accuracy across real-world trajectories.

We evaluate PALMS+ against both PALMS\cite{cheng2024palms} and the recent vision-based baseline F$^3$Loc\cite{chen2024f3loc} on two datasets: the Structured3D dataset~\cite{Zheng2019Structured3DAL}, which primarily contains apartment-like environments, and our own custom-collected dataset of 80 stationary observations across four large campus buildings. To further assess downstream performance, we also test sequential localization on 33 real trajectories using particle filters initialized by heatmaps from all methods. Our experiments show that PALMS+ delivers superior localization accuracy from stationary observations and achieves lower error in sequential tracking, demonstrating its potential as a scalable solution for infrastructure-free indoor localization. To summarize, our main contributions are:
\begin{itemize}
    \item We leverage a depth foundation model to recover geometric structure from monocular images and introduce automatic mechanisms to correct scale inaccuracies.
    \item We introduce PALMS+, a modular system that processes rotating-view images to produce pose posteriors for direct or sequential localization.
    \item We collected and carefully annotated a new dataset that includes panoramas, images, and LiDAR depth from 80 hand-taken observations across 4 large campus buildings. Compared with existing datasets, our custom dataset contains realistic and complex space instances, with substantial geometric ambiguity.
    \item We evaluate PALMS+ on 1) single-observation localization across datasets and 2) sequential localization on 33 real-world trajectories using heatmap-initialized particle filters. We also independently compare our layout matching module against a ray-tracing baseline, showing consistent improvements.

\end{itemize}

\section{Related Work}
\label{sec:related_work}

\subsection{Indoor Localization}
Indoor localization in GPS-denied environments has been approached using a wide range of sensing modalities, including signal-based, retrieval-based, and floor-plan-based approaches. 

Early works in signal-based approaches include Wi-Fi, Bluetooth Low Energy, RFID, Ultra-Wideband, and magnetic field mapping~\cite{Morar2020ACS}. While these methods can yield accurate position estimates, they often rely on prior environment fingerprinting, which requires dense signal measurements and is typically not feasible at scale. Furthermore, maintaining up-to-date signal maps is costly and may not be practical for dynamic or public spaces.

Retrieval-based methods rely on matching the observation with pre-built databases of localized images~\cite{Fan2025MFICNetAM} or point clouds~\cite{Matsumoto2024IndoorVL}. Like signal-based approaches, these methods also suffer from scalability and cost issues.

To address these limitations, recent works have explored floor-plan-based localization \cite{howard2021lalaloc, howard2022lalaloc++, min2022laser, chen2024f3loc, cheng2024palms, li2025flona, Karkus2018ParticleFN, mendez2018sedar, li2020online}, which leverages the structural consistency of walls and corridors typically found in publicly available maps. These methods do not require prior physical access to the building and scale naturally to new environments. Additionally, \cite{li2020online} handles floor plan inaccuracies through stochastic gradient descent and a scale variable. In our approach, we instead blur the layout matching kernel with a Gaussian filter to improve robustness to observation–floor-plan mismatches, and we explicitly test multiple scale candidates during matching.

We categorize floor-plan–based methods by whether they rely on sequential translation data. LaLaLoc++ \cite{howard2022lalaloc++} performs stationary localization from a single panorama by aligning it to the floor plan and selecting the argmax of the resulting heatmap as the pose estimate. F$^3$Loc \cite{chen2024f3loc} extracts one-dimensional depth rays from monocular images and matches them against a pre-computed distance field, producing step-wise heatmaps that are integrated over time with a histogram filter, enabling fully sequential localization from camera inputs. In contrast, PALMS\cite{cheng2024palms} and PALMS+ acquire camera inputs only from a stationary viewpoint, producing heatmaps that serve as priors for sequential localization driven by odometry data. This approach emphasizes infrastructure-free deployment, reduces dependence on continuous video streams, and still enables robust trajectory tracking via a particle filter.

\subsection{Foundation Models and Monocular Depth Estimation}
Recent advances in monocular depth estimation have enabled dense depth prediction from a single RGB image. These approaches are increasingly powered by large-scale foundation models~\cite{Bochkovskiy2024DepthPS, Yang2024DepthAU, Yang2024DepthAV, he2025distill} that learn strong priors from high-quality, diverse datasets. Such models demonstrate strong capability in capturing accurate relative depth across long distances. Among these, we selected Depth Pro~\cite{Bochkovskiy2024DepthPS} for its competitive accuracy and inference speed, replacing LiDAR to generate per-image point clouds from rotational scans (which would be difficult for traditional 3D reconstruction methods due to the lack of camera translation). Lin et al.~\cite{Lin2024PromptingDA} aim to remove scale ambiguity by providing additional low-resolution LiDAR input to one of such models, while our method addresses this limitation using a scale-alignment process, producing consistent geometry suitable for floor plan matching.  

\subsection{Assistive Technologies}
Several smartphone-based systems have been developed to assist blind and visually impaired (BVI) users with indoor navigation. GoodMaps~\cite{goodmapsWorksGoodMaps} and Waymap~\cite{waymapnavWaymapIndoor} rely on visual fingerprinting from prior scans, while others like~\cite{Yoon2019LeveragingAR} use visual-inertial odometry but require the phone to be held up continuously—an approach that is physically demanding and raises usability and safety concerns. Pure IMU-based tracking~\cite{Ren2021SmartphoneBasedIO} and back-tracking methods~\cite{Tsai2024AllTW} enable hands-free use but require an initial position and are fragile to failures. PALMS introduced a “scan-and-walk” paradigm—one stationary scan followed by hands-free tracking—which PALMS+ retains for its user-friendliness and practicality.

\section{Method}
\label{sec:method}
In this section, we first provide an overview of our pipeline and describe our problem formulation. Then, we will share details on the observation module and the layout matching module.

\begin{figure*}
  \centering
  \includegraphics[width=0.8\linewidth]{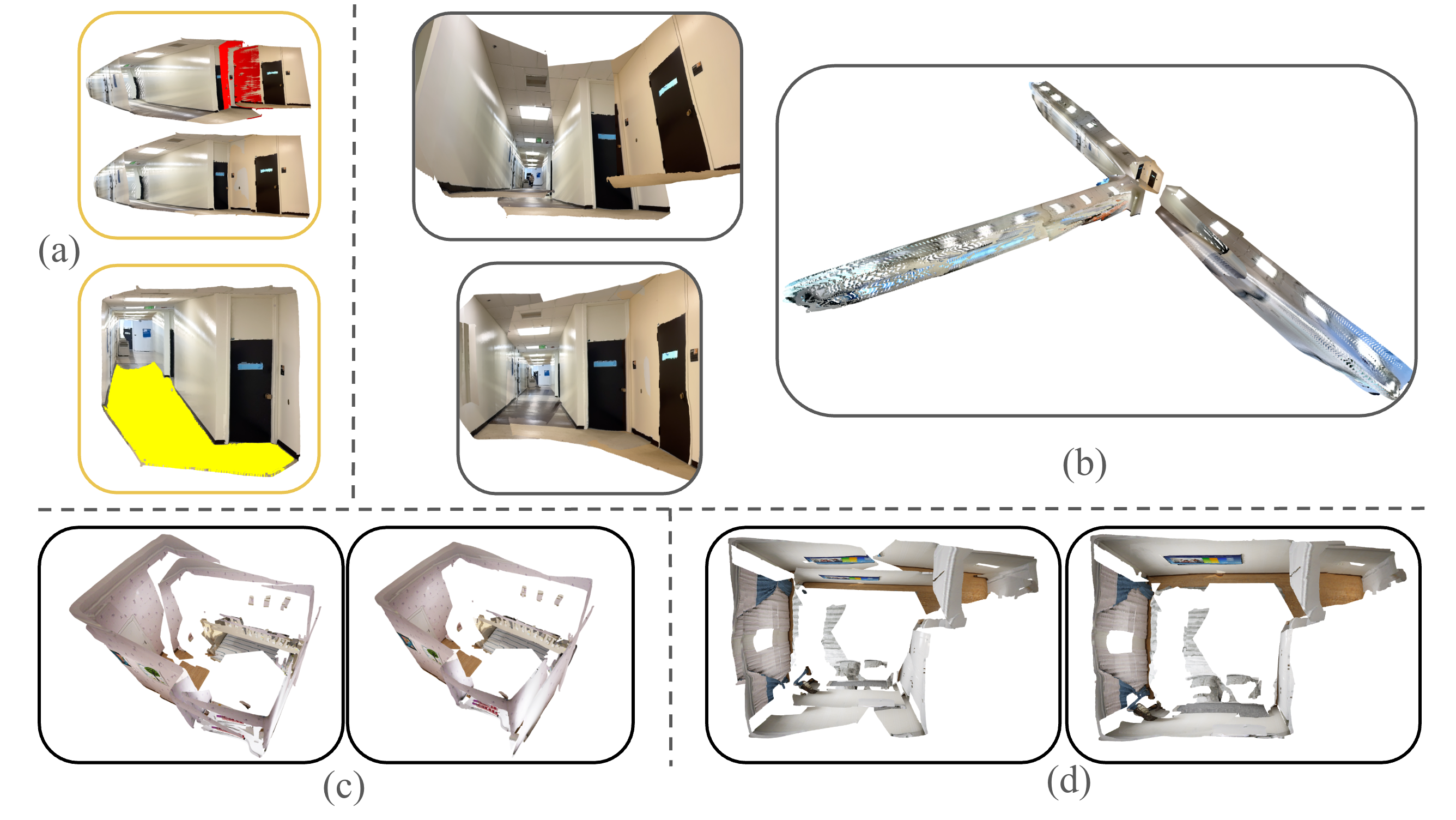}
  \caption{Our point cloud alignment method. (a) Overlap alignment method (top, sub-point clouds used for scale alignment are colored in red) and ground-plane alignment method (bottom, detected ground plane is colored in yellow). (b) An example from our custom dataset, two images on the left show a partial view before and after alignment, and the final aligned point cloud is shown on the right. (c-d) Two examples from the Structured3D dataset, showing the point clouds before (left) and after (right) alignment.}
  \label{fig:pcd_alignment}
\end{figure*}

\subsection{Problem Formulation}
We propose a global indoor localization pipeline that aligns image-derived geometric features with a building's floor plan to estimate the SE(2) pose of the camera relative to the floor plan reference frame, represented as a 2D rigid transformation parameterized by $(x, y, \theta)$. The user performs a stationary horizontal scan by rotating the camera in place, capturing a sequence of images denoted by $\{ I_i \}_{i=1}^{N}$. Relative camera poses between frames are estimated using a pose tracking backend (e.g., Apple's ARKit \cite{ARKit}), and are represented as transformations $\{ \mathbf{T}_{i} \in \mathrm{SE}(3) \}$, where each  $\mathbf{T}_{i}$ defines the pose of image $I_i$  with respect to the reference frame defined by the pose tracking backend.

Each image $I_i$ is processed by a monocular depth estimator to produce a dense depth map $D_i$. Using  $D_i$ , the corresponding pose $\mathbf{T}_i$, and the recorded image intrinsic matrix, we construct local 3D point clouds $ \mathcal{P}_i \subset \mathbb{R}^3 $. These point clouds are transformed into a common frame, passed through a scale alignment module (which corrects for possible scale inconsistencies in the depth maps), and merged to form a global point cloud. 


The aggregated point cloud $\mathcal{P}$ is then projected onto the ground plane, yielding a 2D geometric representation 
$\mathcal{L}_{\text{obs}} = \{l_{obs, 1}, l_{obs, 2}, ..., l_{obs, n} \}$, where each element is a line segment defined by two endpoints in $\mathbb{R}^2$. This 2D projection is then rotated by $O$ candidate angles and matched against the 2D floor plan $\mathcal{F}$ using a scale-invariant geometric matching module, adapted from PALMS. The matching results in $O$ heatmaps $\{ \widetilde{\mathcal{H}}_o \}_{o=1}^{O}$, each representing the likelihood of the observation point over all spatial locations for orientation $\theta_o$. One may take the maximum response across all heatmaps to obtain the predicted observation pose as
\begin{equation}
    (x^*, y^*, \theta^*_o) = \arg\max_{(x, y, o)} \widetilde{\mathcal{H}}_{o}(x, y)
    \label{eq:1}
\end{equation}

The above pipeline is separated into the \textbf{observation module} that produces $\mathcal{L}_{\text{obs}}$, and the \textbf{layout matching module} that produces the heatmaps $\{ \widetilde{\mathcal{H}}_o \}_{o=1}^{O}$. 

\subsection{The Observation Module}
\label{sec:observation}
\textbf{Point cloud reconstruction from observation.} Given images $I_i$ taken in a scan from a certain location, we compute dense depth maps $D_i$ and, from the camera intrinsics,  produce local point clouds $\mathcal{P}_i$ as described above. Each $\mathcal{P}_i$ is transformed into a common reference frame using $\mathbf{T}_i$. Since the depth maps $D_i$ are estimated from a monocular image, they may be affected by a scale error, in particular when the scene is very close or in the absence of sufficient context. This error typically varies from view to view. 

We estimate relative scales (with respect to a chosen reference view) by computing $N-1$ scale factors that maximize overlap between neighboring point clouds. Specifically, for each point in the $i$-th cloud, we find its nearest neighbor in the overlapping cloud, and jointly optimize the set of scale factors to minimize these distances (with $i=1$ as the reference view):

Formally, we jointly optimize the set of scale factors as:
\begin{equation}
    \{\lambda_i\}_{i=2}^N = \arg\min_{\{\lambda_i\}} \sum_{(m, n)} d\bigl(\lambda_m \cdot \mathcal{P}_m' ,  \lambda_n\cdot\mathcal{P}_n'\bigr)
    \label{eq:2}
\end{equation}
where $d\bigl(P, Q\bigr)$ denotes the weighted nearest-neighbor distance between a pair of neighboring point clouds:
\begin{equation}
    d\bigl(P, Q\bigr) = \frac{1}{|P|} \sum_{p \in P} \frac{\|p - \text{NN}_Q(p)\|}{\|p - \text{NN}_Q(p)\|_{[0,1]} + 1}
    \label{eq:3}
\end{equation}
Here, $\text{NN}_Q(p)$ is the nearest neighbor of $p$ in $Q$, and $||\cdot||_{[0,1]}$ normalizes the nearest neighbor distances (for all $p\in P$) between 0 and 1. In practice, this expression robustifies the residual by giving lower weight to larger distances. We evaluated several robust normalization strategies and found this one to be effective.
The minimization in \cref{eq:2} is computed using L‑BFGS‑B~\cite{Byrd1995ALM}, an iterative quasi-Newton method with limited memory and bound constraints.
The scale factors $\{\lambda_i\}$ thus found are used to globally adjust the scales of all local point clouds. Once aligned, the scaled point clouds are merged to produce a single scale-corrected global point cloud $\mathcal{P}=\bigcup_{i}\lambda_i\mathcal{P}_i$. 
Note however that $\mathcal{P}$ may still be affected by global scale error. To mitigate this, we fit a ground plane to the points in $\mathcal{P}$ that are classified as being on the ground (using RANSAC plane detection), and adjust the global scale by assigning a canonical camera height of 1.5~m above this ground plane. We show qualitative results for this alignment process in \cref{fig:pcd_alignment}. Additional details are provided in the supplementary material.


\noindent \textbf{Point cloud projection and segment extraction.}
To enable geometric matching with the floor plan, $\mathcal{P}$ is projected onto the horizontal plane. Specifically, we extract all points within a vertical band of $\pm 0.1$ meters around the camera height,
and discard the vertical coordinate.
This results in a 2D point set $\mathcal{P}_{\text{2D}} \subset \mathbb{R}^2$ that captures the horizontal structure of the scene.
To convert $\mathcal{P}_{\text{2D}}$ into a representation suitable for geometric matching, we rasterize the point set and apply Canny edge detection~\cite{Canny1986ACA} followed by a probabilistic Hough transform~\cite{Matas2000RobustDO}. The resulting line segments $\mathcal{L}_{\text{obs}}$ capture dominant structural boundaries such as walls and doorways—features commonly found in floor plans. These extracted segments serve as the geometric observation used in the subsequent matching stage.

\begin{figure*}
  \centering
  \includegraphics[width=0.8\linewidth]{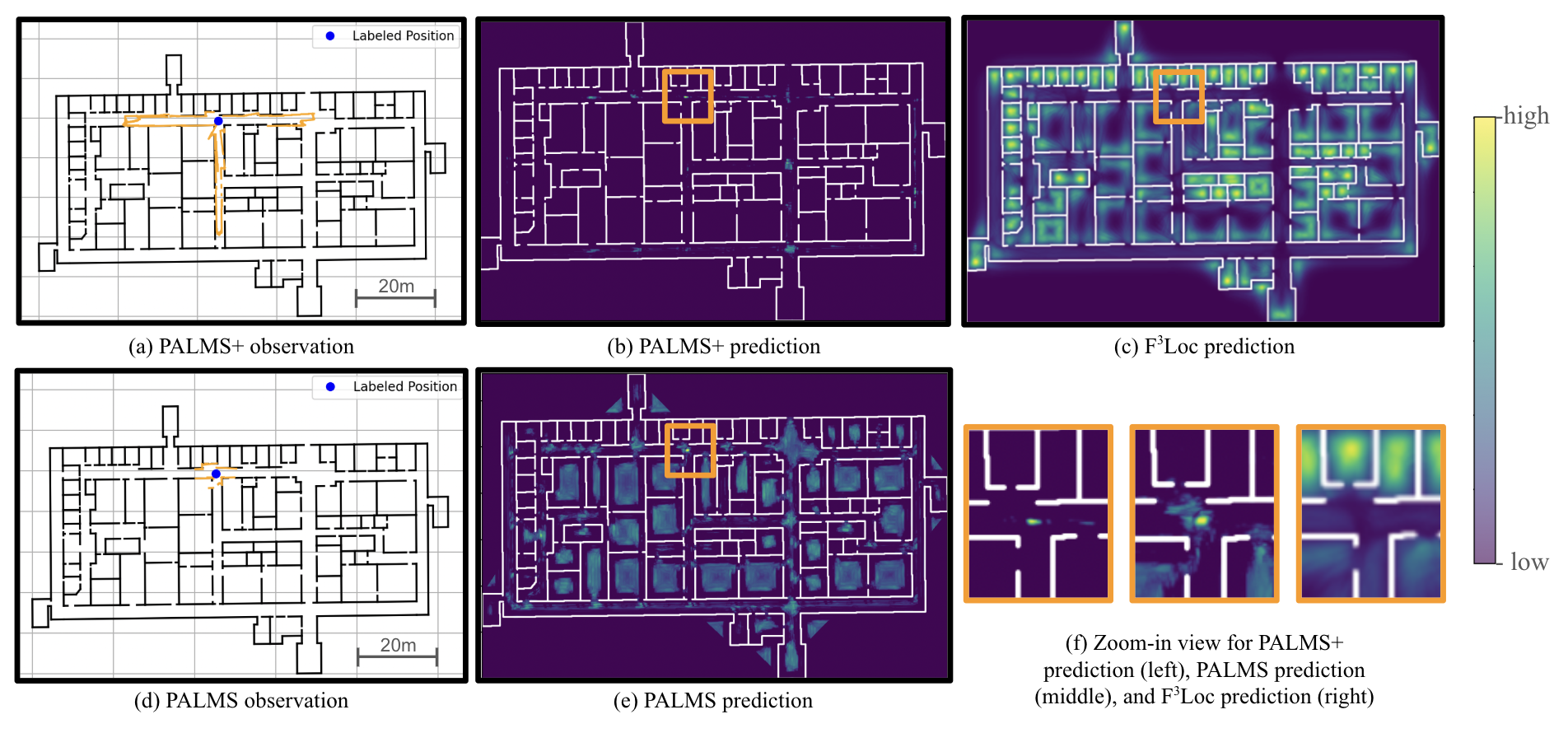}
  \caption{Qualitative analysis. We removed the orientation dimension from the heatmaps by taking the max. Each heatmap is then normalized to a PDF. (a) PALMS+'s projected geometry from the reconstructed point cloud. (b) PALMS+ heatmap output. (c) F$^3$Loc heatmap output. (d) PALMS's projected vertical planes extracted by ARKit LiDAR. (e) PALMS heatmap output. (f) Zoomed-in view for the heatmap around the ground-truth location for all three methods. The green circle on the heatmaps indicates the ground-truth location.}
  \label{fig:palms+_exp}
\end{figure*}

\subsection{The Layout Matching Module}
\label{sec:layout_matching}
The process for generating localization heatmaps in PALMS+ follows the same principles used for PALMS~\cite{cheng2024palms}, with an extension to incorporate scale uncertainty and a new way to generate orientation candidates. As in PALMS, the objective is find a location and orientation in the floor plan that are consistent with the observed geometric features. 

Ideally, the recorded walls (line segments $\mathcal{L}_{obs}$) should align with the corresponding walls in the floor plan when placed at the correct location and orientation. For a given pose, this alignment can be quantified by the inner product of the two bitmaps. To evaluate all candidate locations at a fixed orientation $\theta_o$, we convolve the floor plan with a kernel derived from the rotated recorded-wall bitmap ($RW_{\theta_o}$). Robustness is improved by Gaussian-smoothing $RW_{\theta_o}$ and incorporating a rotated {\em Certainly Empty Space} ($CES_{\theta_o}$) kernel to enforce visibility constraints. As shown in the “Kernel Creation” block in \cref{fig:palms+}, the $RW_{\theta_o}$ is created from the observed wall segments, and the $CES_{\theta_o}$ region corresponds to the triangles formed by the observation point and each observed wall segments; it enforces that no other wall in the floor plan should appear inside these triangles. The final heatmap $\mathcal{H}_o$ is obtained by convolving the floor plan with $RW_{\theta_o} - \alpha CES_{\theta_o}$ (with $\alpha = 10$ for our experiments).





\noindent\textbf{Extracting candidate orientations.} 
Rather than testing a dense set of orientations as in F$^3$Loc, or assuming a Manhattan world as in PALMS, we identify a small set of candidate orientations without constraining walls to orthogonal layouts. We assume only that walls are aligned with a limited set of dominant directions. Accordingly, the histogram of wall segment orientations in the floor plan contains a few distinctive peaks. For a correctly oriented observation, the histogram of observed line directions $\mathcal{L}_{\text{obs}}$ should align with these peaks. When the viewpoint is rotated by angle $\theta_o$, the histogram of observed directions undergoes a circular shift. Candidate orientations are therefore obtained by finding the shifts that maximize alignment between the floor-plan and observation histograms. This is computed efficiently via circular convolution, and the top $O/2$ peaks are selected, with their 180$^\circ$ complements giving the other $O/2$ orientations. We use $O=10$ in stationary localization and $O=4$ in sequential localization.


\noindent \textbf{Accounting for scale ambiguity.} Although cross-view scale consistency is enforced by the algorithm of \ref{sec:observation}, the whole point cloud may be affected by a residual global scale error, which could negatively influence the layout matching operation. To address this, we consider a small set  $S$ of corrective scale factors which are applied to the point cloud, and compute heatmaps $\mathcal{H}_{o, s}, s\in S$ ($S=\{0.9, 1.0, 1.1\}$ in our experiments). In principle, at the ``right" corrective factor, the observed lines $\mathcal{L}_{\text{obs}}$ should best match the lines in the floor plan, resulting in a high value of the heatmap. Accordingly, we marginalize out the scale correction terms by simply taking the maximum over $s$ of $\mathcal{H}_{o, s}$ for each spatial location and orientation.

\section{Experiments}
\label{sec:experiments}

\subsection{Datasets}
\label{sec:palms+_datasets}

\textbf{Structured3D dataset.}
We used the official validation and testing splits of the Structured3D (S3D) dataset~\cite{Zheng2019Structured3DAL}, which comprises 500 scenes. Specifically, we sampled from the panorama subset with full furniture and raw lighting configurations using code from~\cite{min2022laser}. When extracting images from the panoramas, we sampled $640 \times 480$ images with a horizontal FOV of $108^\circ$, consistent with the training setup of F$^3$Loc. 
To ensure the ground plane is visible (given the relatively small spaces in S3D) we applied a pitch of $-15^\circ$ during sampling.

\noindent
\textbf{Custom dataset.}
PALMS+ targets indoor localization in large public spaces, where its long-range observations help resolve geometric ambiguities—an advantage less applicable in smaller environments. Because no existing dataset fits this setting, we collected a custom dataset using an iPhone 14 Pro and a custom ARKit-based app. Each observation involved a hand-held, portrait-mode 360° scan, with snapshots captured approximately every 20°, averaging 18.2 images per scan. ARKit provided visual-inertial pose tracking throughout. Each snapshot includes:
\begin{itemize}
    \item An RGB image of size $1920 \times 1440$.
    \item A depth map from the iPhone's LiDAR sensor at $256 \times 192$ resolution, with associated depth confidence map
    \item Detected vertical planes via ARKit.
    \item Camera intrinsics and pose information.
\end{itemize}
Each image spans a horizontal field of view (FOV) of approximately $60^\circ$. The dataset encompasses 80 observations across 4 large campus buildings, totaling 1,456 images. The distribution of locations and examples can be found in the supplementary materials. Additionally, we collected one panorama for each observation using iPhone 14 Pro's ultra-wide camera and built-in panorama capturing algorithm. This allowed us to run direct comparisons with F$^3$Loc.
Each panorama was manually processed to be in equirectangular format, from which we sampled perspective images using the same sampling code from~\cite{min2022laser}.

\noindent
\textbf{Sampling strategy.}
For the S3D dataset, we sampled images at $60^\circ$ intervals over a full $360^\circ$ rotation, resulting in 6 images per observation. For the custom dataset, the panoramas only cover about $320^\circ$, so only 5 images per observation are evenly sampled. To provide insights on the number of images needed to produce accurate localization, we evaluated our methods under three distinct settings:
\begin{enumerate}
    \item \textbf{Full-view:} Utilizes all 6 (or 5) images, providing comprehensive environmental coverage.
    \item \textbf{Partial-view:} Employs 3 consecutive images, covering approximately $228^\circ$ ($108^\circ +2\times60^\circ$).
    \item \textbf{Single-view:} Uses only one image, representing an extreme case with minimal observational data.
\end{enumerate}
For the partial-view and single-view settings, we first generated a point cloud for each of the images in an observation and performed ground plane detection on each. We then identified the image with the highest number of pixels on the ground plane; this image was selected as the central image for the partial-view setting and as the sole image for the single-view setting. For PALMS, we cropped the detected planes using the corresponding FOV in each setting.

\textbf{Handling large depth artifacts}. Depth Pro sometimes captures geometry beyond the intended space through transparent surfaces like windows or glass doors, which disrupts scale alignment and layout matching. To mitigate this, we also tested with manually masking such regions in our custom dataset and by excluding windows in the S3D dataset (denoted by PALMS+$^*$). 

\subsection{Full System Experiment}
\label{sec:full_system_exp}
\begin{table}[h]
  \centering
  {
    
  \small
  \renewcommand{\arraystretch}{0.8}
  \begin{tabular}{l|l|ccccc}
    \toprule
     \multirow{2}{*}{\textbf{Setting}} &\multirow{2}{*}{\textbf{Method}} & \multicolumn{5}{c}{Loc. Acc. @1m / 1m~30$^\circ$ (\%) $\uparrow$} \\
     \cmidrule(lr){3-7}
    & & 1 & 2 & 3 & 4 & All\\
     
    \midrule
    \multirow{4}{*}{Full} & F$^3$Loc & 0.0  &0.0  & 0.0 & 0.0 & 0.0 \\
    &PALMS  & 26.3 & 0.0 & 0.0 & 5.6 & 7.6\\
    &PALMS+ & 52.6 & 8.3 & 33.3 & 33.3 & 30.4 \\
    &PALMS+$^*$ &  \textbf{57.9} &  \textbf{12.5} & \textbf{50.0} & \textbf{38.9} & \textbf{38.0} \\
    \midrule
    \multirow{4}{*}{Partial} & F$^3$Loc & 0.0  &0.0  & 0.0 & 0.0 & 0.0 \\
    &PALMS  & 10.5 & 0.0 & 0.0 & 11.1 & 5.1\\
    &PALMS+ & 31.6 & \textbf{12.5} & \textbf{11.1} & 22.2 & 19.0 \\
    &PALMS+$^*$ & \textbf{36.8} & \textbf{12.5}  & \textbf{11.1} & \textbf{27.8} & \textbf{21.6} \\
    \midrule
    \multirow{4}{*}{Single} & F$^3$Loc & 0.0  &0.0  & 0.0 & 0.0 & 0.0 \\
    &PALMS  & 5.3 & 0.0 & 0.0 & 0.0 & 1.3 \\
    &PALMS+ & 15.8 & 4.2  & \textbf{5.6} & \textbf{22.2} & 11.6\\
    &PALMS+$^*$ & \textbf{21.1} & \textbf{8.3} & \textbf{5.6} & \textbf{22.2} & \textbf{14.0} \\
    \bottomrule
  \end{tabular}
  \caption{Localization accuracy on the custom dataset in different buildings (represented by 1$\sim$4 on the header, ``All" is the result for all buildings). PALMS+$^*$ is PALMS+ with masked depths. Localization accuracy @1m and @1m$30^\circ$ share the same values.}
  \label{tab:custom_result}
  }
\end{table}

\begin{table}[h]
  \centering
  \small
  \renewcommand{\arraystretch}{0.8}
  \begin{tabular}{l|ccc}
    \toprule
     \multirow{2}{*}{\textbf{Method}} & \multicolumn{3}{c}{Loc. Acc. @1m / 1m~30$^\circ$ (\%) $\uparrow$} \\
     \cmidrule(lr){2-4}
     & Full & Partial & Single \\
    \midrule
     F$^3$Loc & 8.2 / 1.2  &  6.7 / 1.3& 6.5 / 1.4 \\
     PALMS+ & 17.9 / 8.9  & 11.5 / 5.3   & \textbf{8.6} / \textbf{3.6}  \\
     PALMS+$^*$ & \textbf{19.3} / \textbf{10.7} & \textbf{12.5} / \textbf{6.2}  & 8.0 / 3.4 \\
    \bottomrule
  \end{tabular}
  \caption{Localization accuracy on the S3D dataset under different observation settings. PALMS+$^*$ is PALMS+ with masked depths.}
  \label{tab:s3d_result}
\end{table}

\noindent
\textbf{Baselines.}
We compare our method against two baselines: the original PALMS algorithm and the recently proposed F$^3$Loc~\cite{chen2024f3loc}. F$^3$Loc is conceptually similar to our approach -- It takes a sequence of perspective images as input, performs monocular depth estimation, and uses a layout matching algorithm to compute a probability distribution over all floor plan locations and orientation candidates. 

Although F$^3$Loc supports both single-image and multi-view depth estimation, its multi-view module is specifically designed to leverage translation-induced parallax to improve monocular depth quality—an advantage that does not apply in our scenario, which involves little translation in scans. For this reason, we restrict our comparison to the monocular depth estimation module of F$^3$Loc. The provided weights were trained on their iGibson dataset, because the S3D-trained weights are not available at the time of this writing. 

For PALMS, which requires vertical plane detection using the iPhone’s LiDAR sensor, we limit its evaluation to the custom dataset, where such data is available. Although the original PALMS only considers 4 principal orientations, we  followed F$^3$Loc's approach and considered 36 orientation candidates for PALMS. 

\noindent
\textbf{Evaluation metrics.}
To evaluate localization accuracy, we report the following metrics: \textbf{Localization accuracy @1m / @1m~30$^\circ$} -- The proportion of predictions within 1 meter / 1 meter and 30 degrees of the ground-truth location. These metrics are computed across all samples in the evaluation set. We also put additional results in the supplementary.

\noindent
\textbf{Quantitative evaluation.} Localization accuracies are reported in \cref{tab:custom_result} for each building in our custom dataset, and \cref{tab:s3d_result} for all indoor scenes in the S3D dataset. We see that PALMS+ outperforms both baselines by significant margins, especially on our custom dataset, where PALMS+ achieves a pose estimation accuracy @1m$30^\circ$ of 30.4\% across all four buildings (38.0\% with depth masks). As expected, pose accuracy decreases with decreasing FOV ({\em partial} and {\em single} settings).
We note that PALMS+'s performance (\textit{full}, \textit{masked} Acc.@1m~30$\%$) varies noticeably across buildings (ranging from $57.9\%$ to $12.5\%$), with lower accuracy for buildings containing repeating patterns (building 2).
Remarkably, the pose returned by F$^3$Loc was always more than 1~m away from the ground truth for our custom dataset, though it was within 5~m for 4\% of the time ({\em single} setting, see supplementary material).

\begin{figure}[t]
  \centering
   \includegraphics[width=1\linewidth]{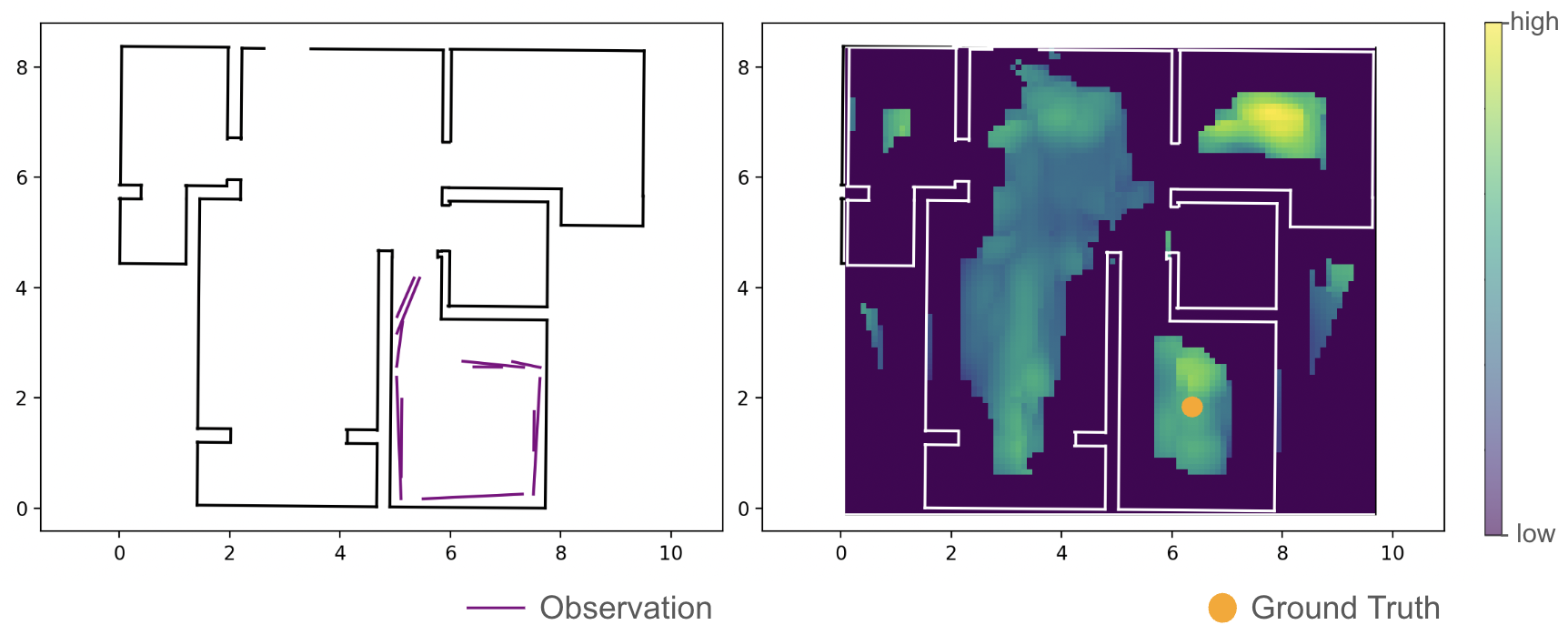}
   \caption{An example from the S3D dataset (same as \cref{fig:pcd_alignment}d), showing the observed line segments (left), and the ambiguous posterior distribution (right), which contains an area of high probability in another room away from the ground-truth.}
   \label{fig:s3d_example}
\end{figure}

PALMS+ performs worse on the S3D dataset because most observations are from small rooms with homogeneous geometry, which often yield multimodal posteriors (see \cref{fig:s3d_example}). Contrarily, our custom dataset emphasizes transitional spaces (hallways, intersections, lobbies) that provide richer geometric cues. In \cref{tab:s3d_result}, we also note a slight drop in single-view performance with masked depths, likely due to the loss of localization-relevant information.



\noindent
\textbf{Qualitative evaluation.} We show  sample heatmaps from a single observation for PALMS+, PALMS, and F$^3$Loc in \cref{fig:palms+_exp}. We observe that the heatmap modes produced by F$^3$Loc often have large values inside rooms, likely due to the model being trained primarily on small indoor spaces, which limits its depth predictions beyond a certain range. This would explain its low localization accuracy on our custom dataset. 
In contrast, PALMS and PALMS+ leverage distinctive local geometry to infer location, with PALMS+ producing noticeably more concentrated heatmaps. While PALMS+ benefits from a longer observation range ($\approx$25m) that helps reduce geometric ambiguity, it can still struggle in homogeneous areas—such as long hallways—where the resulting heatmaps tend to exhibit elongated modes rather than sharp peaks (see supplementary material).

\subsection{Sequential Localization}
To showcase how our heatmaps can be used for downstream sequential localization for pose refinement and camera-free tracking, we collected 33 trajectories, each with more than 100 steps, from 3 buildings where we also collected the custom dataset, and performed Bayesian accumulation using a particle filter. The number of trajectories for buildings $1 \sim 3$ are $[8, 19, 6]$, respectively. We follow the same approach as in \cite{cheng2024palms} to collect odometry data using ARKit, and estimate the ground truth trajectory using a particle filter initialized with a known pose. 

\begin{table}[h]
  \centering
  \small
  \renewcommand{\arraystretch}{0.8}
  \begin{tabular}{l|l|cccc}
    \toprule
     \multirow{2}{*}{\textbf{Metric}} &\multirow{2}{*}{\textbf{Method}} & \multicolumn{3}{c}{\textbf{Building \#}} \\
     \cmidrule(lr){3-6}
     & & 1 & 2 & 3 & All \\
     
    \midrule
    \multirow{4}{*}{RMSE (m) $\downarrow$} & F$^3$Loc & 50.5  &  10.7 & 9.5 &  17.6 \\
    & PALMS  & 38.1 & 2.6 & 3.4 &  4.1 \\
    & PALMS+ & 2.1  &  3.7 &  2.1 & 2.2 \\
    & PALMS+$^*$ & \textbf{0.8} & \textbf{1.7}  & \textbf{1.6} & \textbf{1.6} \\
    \midrule
    \multirow{4}{*}{Loc. Err. (m) $\downarrow$} & F$^3$Loc & 52.3 & 9.5  & 10.3 & 18.3 \\
    & PALMS  & 0.9 & 2.4 & 3.1 &  2.1 \\
    & PALMS+ & 0.6 & 3.7 & \textbf{1.6} & 2.0 \\
    & PALMS+$^*$ & \textbf{0.4} & \textbf{1.4}  & \textbf{1.6} & \textbf{1.3 }\\
    \bottomrule
  \end{tabular}
  \caption{Sequential localization accuracy in different buildings. PALMS+$^*$ is PALMS+ with masked depths.}
  \label{tab:seq_loc}
\end{table}

Following F$^3$Loc\cite{chen2024f3loc}, we measure the Root Mean Squared Error (RMSE) of the last 10 steps and the localization error at the last step. The median values of our results are shown in \cref{tab:seq_loc}. We can see that PALMS+ and PALMS+$^*$ outperformed the other methods. For building 1, PALMS shows an irregularly high contrast between its RMSE and localization error. By visualizing the particles, we found that this is caused by the particle filter converging at the last few steps from a bimodal distribution \cite{Ren2021SmartphoneBasedIO}. 

\begin{figure}[t]
  \centering
   \includegraphics[width=0.8\linewidth]{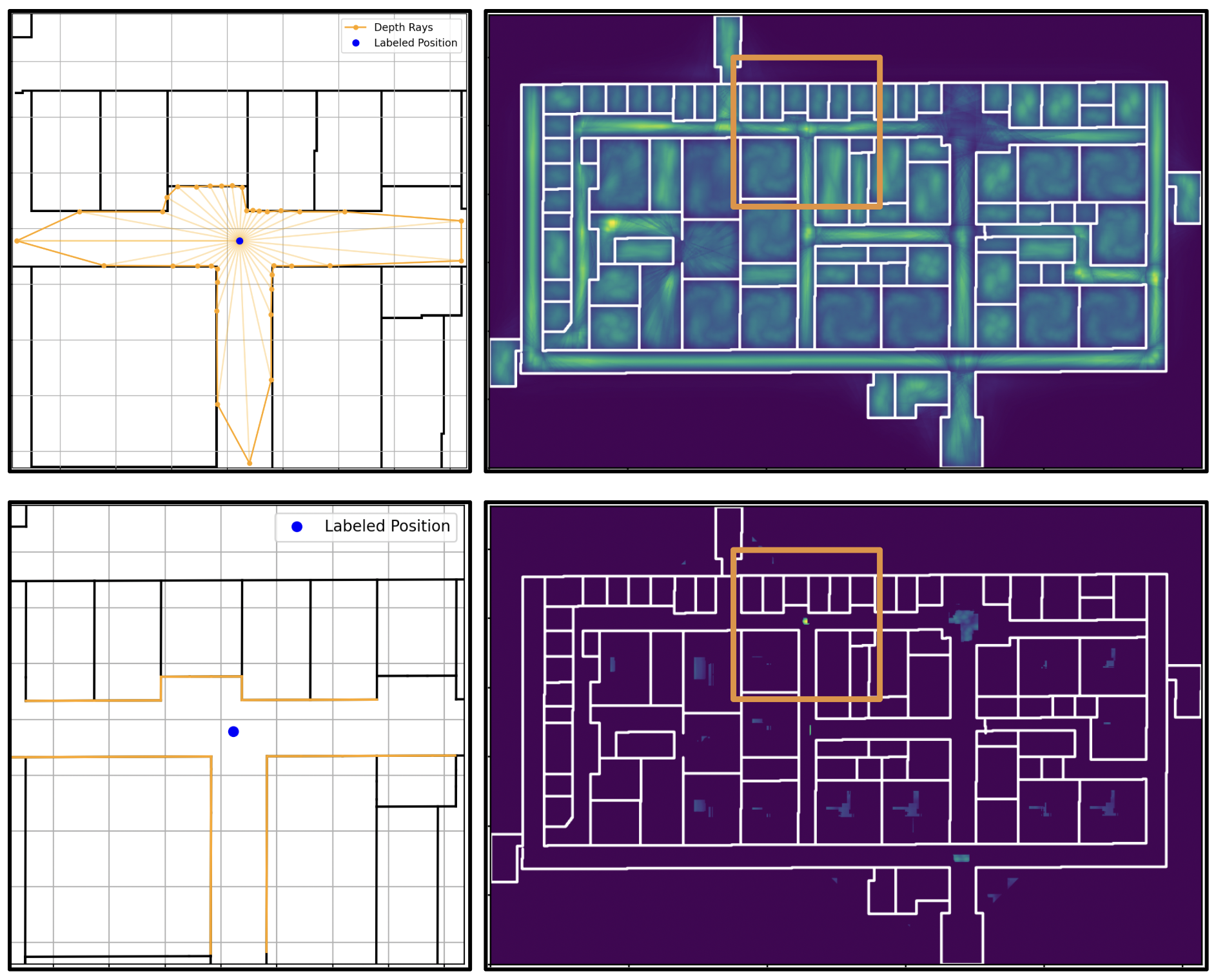}
   \caption{An example of the ``perfect observations'' sampled from the floor plan (left), and heatmaps obtained using corresponding layout matching methods (right). Top: F$^3$Loc. Bottom: PALMS+. The orange bounding boxes outline the same regions on the left.}
   \label{fig:layout_matching}
\end{figure}

\subsection{Layout Matching Module Evaluation}
The performance of the overall localization method depends largely on the chosen layout matching algorithm. Here we compare two layout matching algorithms  under ideal observation conditions: PALMS' kernel-based method\cite{cheng2024palms}, briefly summarized in~\cref{sec:layout_matching}; and F$^3$Loc's ray-tracing method, using a set of 36 equiangular rays.  Specifically, we use the floor plan to measure the distance, from a certain location, to all wall surfaces visible from that location. This provides an error-free proxy for the observation module. The maximum observation distance was set to be 10~m for both methods, and all doors were manually ``closed". An example is shown in \cref{fig:layout_matching}. For each considered location, we computed heatmaps with each method, then selected the mode (location of the maximum) for each heatmap.  On our custom dataset, we measured a localization accuracy @1m~$30^\circ$ of $11.2\%$ for the ray-tracing method and $88.7\%$ for the kernel-based method. This result suggests that when accurate, high-resolution observations are available, the kernel-based layout matching method can provide higher localization accuracy than the ray-tracing method. Increasing the number of rays to 180 slightly improves the ray-tracing results by around 3\%.

\subsection{Ablation Study}
To assess the impact of scale alignment, we conduct an ablation study by restricting how scales are estimated and compare against our \textit{Full} algorithm. In the \textit{Ground} setting, scales are estimated solely from detected ground planes, and only point clouds with valid ground planes are retained. In the \textit{None} setting, no scale alignment is applied. Using full-view data sampled from our custom dataset, we aggregate results across all buildings (\cref{tab:ablation}). The results show that combining overlap-based alignment with ground-plane estimation yields a substantial performance gain.

\begin{table}[ht]
  \centering
  \small
  \renewcommand{\arraystretch}{0.8}
  \begin{tabular}{c|cc}
    \toprule
     \textbf{Scale} & \multicolumn{2}{c}{Loc. Acc. @1m / 1m~30$^\circ$ (\%) $\uparrow$} \\
     \cmidrule(lr){2-3}
     \textbf{Alignment}& PALMS+ & PALMS+$^*$ \\
    \midrule
     \textit{None} & 10.0 / 8.9 & 12.4 / 10.0\\
     \textit{Ground} & 20.8 / 20.8 & 26.4 / 25.1 \\
     \textit{Full} & \textbf{30.4} / \textbf{30.4} & \textbf{38.0} / \textbf{38.0} \\
    
    \bottomrule
  \end{tabular}
  \caption{Ablation study on the contribution of the scale alignment algorithm. PALMS+$^*$ is PALMS+ with masked depths.}
  \label{tab:ablation}
\end{table}

\section{Conclusion}
\label{sec:conclusion}
We introduced PALMS+, a modular, image-based indoor localization system that reconstructs 3D geometry from rotational scans, aligns it with a floor plan through kernel-based layout matching, and produces interpretable heatmaps over position and orientation. These heatmaps enable both direct localization and sequential tracking when combined with odometry, making PALMS+ a practical solution for smartphone-based indoor navigation.
Experiments on Structured3D and our custom dataset show that PALMS+ consistently outperforms PALMS and F$^3$Loc in stationary localization as well as sequential localization via particle filtering.



\section*{Acknowledgements}
This research was funded in part by the National Eye Institute, National Institutes of Health, under grant number R01EY036360. The authors would like to thank Loni Halsted-Ruelas for her invaluable assistance in dataset curation and experimental support.
{
    \small
    \bibliographystyle{ieeenat_fullname}
    \bibliography{main}
}

\clearpage
\appendix
\onecolumn
\section*{Supplementary Material for\\
PALMS+: Modular Image-Based Floor Plan Localization Leveraging Depth Foundation Model}
\noindent\textit{Yunqian Cheng \quad Benjamin Princen \quad Roberto Manduchi}\\
University of California, Santa Cruz, United States \\
{\tt\small \{ychen827, bprincen, manduchi\}@ucsc.edu}

\section{Technical Details}

\subsection{Layout Matching}
\begin{figure}[h]
    \centering
    \resizebox{0.5\textwidth}{!}{\includegraphics[width=\linewidth]{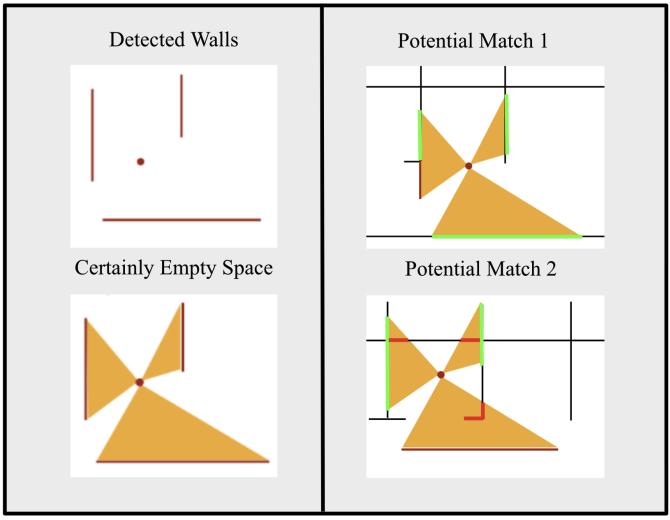}}
    \caption{Illustration of the Certainly Empty Space (CES) from \cite{cheng2024palms}. It shows a potentially good match (top right) and a bad match (bottom right) because of the floor plan segments' intersection with the CES, which breaks the visibility constraint.}
    \label{fig:CES}
\end{figure}

Given observed wall segments, we employ a convolution-based approach to estimate the likelihood that a particular location corresponds to the current position. The convolution kernel is constructed by rasterizing the observed wall segments. To account for uncertainty in the orientation, the vector map of wall segments is rotated $O$ times, yielding $O$ distinct kernels.

\begin{figure}[t]
    \centering
    \resizebox{0.8\textwidth}{!}{\includegraphics[width=\linewidth]{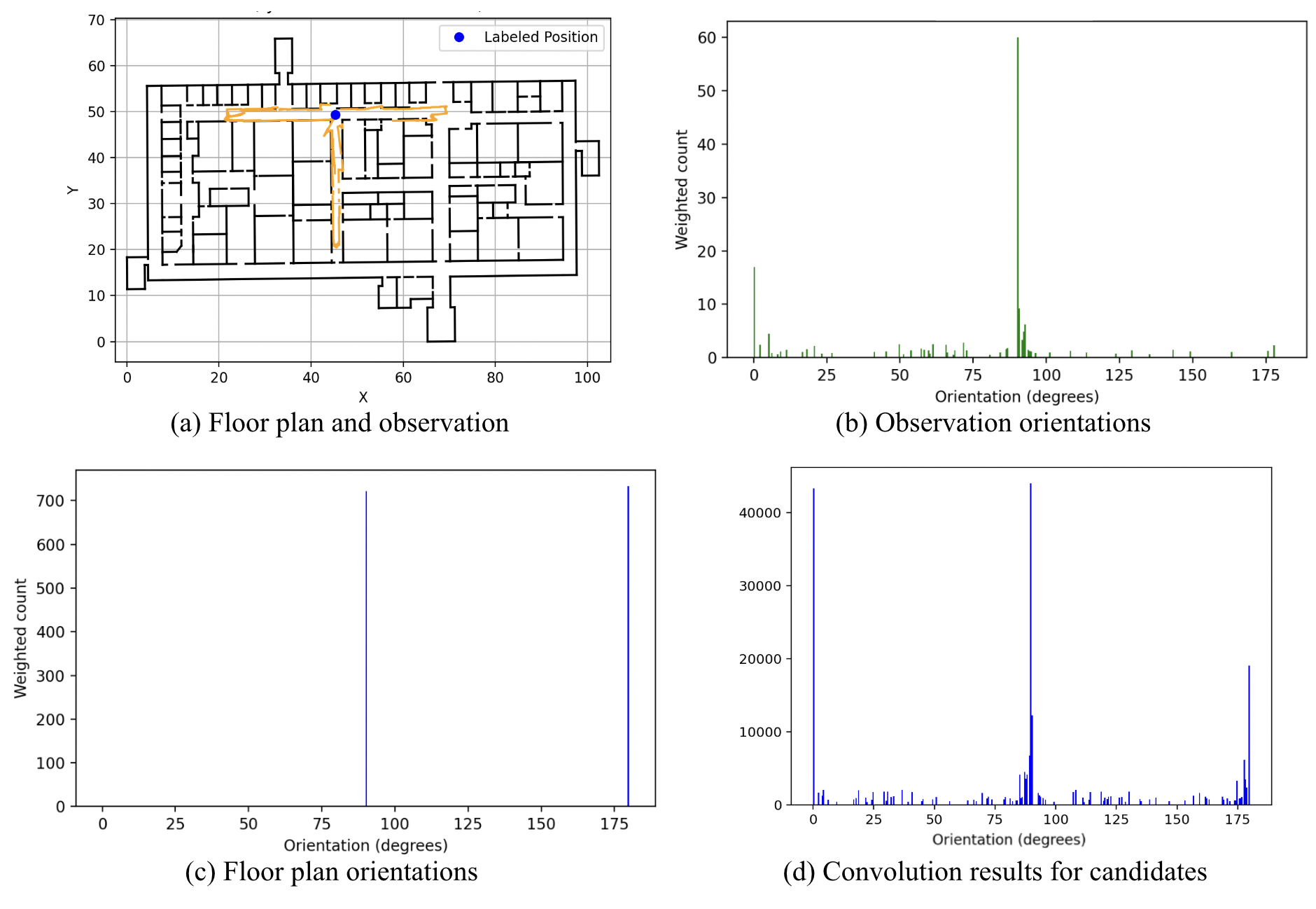}}
    \caption{Illustration of the candidate orientation extraction process. For each set of line segments, we extract the orientations and weight each with the segment length.}
    \label{fig:top_ori}
\end{figure}

The orientations of wall segments are extracted directly from their vector form and normalized to the interval $[-\pi/2, \pi/2]$. The same process is done on the vector floor plan, then we find the orientation candidates using the method illustrated in \autoref{fig:top_ori}. Let $\Theta = \{\theta_1, \theta_2, \dots, \theta_5\}$ denote the top five estimated orientation candidates. Because wall segment orientations are ambiguous up to $\pi$ (i.e., a vector direction cannot distinguish between $\theta$ and $\theta + \pi$), we augment this set by adding $\pi$ to each element of $\Theta$, producing
\[
\Theta' = \{\theta_1, \dots, \theta_5, \theta_1+\pi, \dots, \theta_5+\pi\}.
\]

The convolution kernels are then constructed by rotating the vector map according to each $\theta \in \Theta'$, resulting in $O = 10$ distinct kernels. The corresponding convolution operations yield spatial distributions of likelihood values across the floor plan.

However, this baseline approach does not inherently enforce visual occlusion constraints. If a wall is observed, no additional walls should be detected between the observer and that wall. This principle, termed \textit{Certainly Empty Space} (CES), leverages the geometry of observed walls to define regions that are guaranteed to remain unoccupied. Concretely, the triangular region subtended by an observed wall segment and the observer’s position defines an exclusion zone within which no additional wall segments should appear. Each exclusion zone is defined by three points: the observer’s location and the two endpoints of the wall segment. These exclusion zones, together with the vector map, are rotated according to the orientations in $\Theta'$ and rasterized to obtain the final set of convolution kernels. CES is illustrated in \autoref{fig:CES}.

\subsection{Runtime Analysis}
We measured the runtime of each component in our pipeline, from point cloud reconstruction to the final posterior computation, on a device equipped with an Apple M2 chip. The point cloud reconstruction and scale alignment steps took 20.447 seconds, while projecting the point cloud to line segments and determining their candidate orientations each required 0.005 seconds. The final posterior estimation took 3.773 seconds.

Overall, the current implementation is not yet suitable for real-time deployment. However, there is substantial room for optimization. The runtime could be reduced through code-level optimization, lowering image or point cloud resolution, or leveraging GPU acceleration. We exclude the runtime of the depth foundation model, as its performance metrics are reported in the original paper.

\section{Additional Experimental Results}
We show additional experimental results from PALMS~\cite{cheng2024palms} and F$^3$Loc~\cite{chen2024f3loc} as well as our own results. These results, as well as results from Tables \ref{tab:custom_result} and \ref{tab:s3d_result} of the main paper, are from the same experiments.

We note that for the custom dataset, the localization accuracy is the same for @1m and @1m30$^\circ$. This is expected because our dataset emphasizes transitional spaces—hallways, intersections, and lobbies—that provide richer geometric and directional cues. As a result, when the position estimate is accurate (within 1 m), the inferred orientation is almost always consistent (within 30$^\circ$) as well.

\begin{table}[h]
  \centering
  \small
  \resizebox{0.5\textwidth}{!}{
  \renewcommand{\arraystretch}{1.2}
  \begin{tabular}{l|l|cccccc}
    \toprule
     \multirow{2}{*}{\textbf{Building}} & \multirow{2}{*}{\textbf{Method}} & \multicolumn{6}{c}{Localization Accuracy (\%)} \\
     \cmidrule(lr){3-8}
     & & 0.1m & 0.5m & 1m & 1m $30^\circ$ & 2m & 5m \\
    \midrule
    \multirow{4}{*}{1}
      & F$^3$Loc   & 0.0 & 0.0 & 0.0 & 0.0 & 0.0 & 0.0 \\
      & PALMS      & 15.8 & 26.3 & 26.3 & 26.3 & 26.3 & 26.3 \\
      & PALMS+     & 0.0 & 47.4 & 52.6 & 52.6 & 57.9 & 57.9 \\
      & PALMS+$^*$ & 0.0 & 52.6 & 57.9 & 57.9 & 68.4 & 68.4 \\
    \midrule
    \multirow{4}{*}{2}
      & F$^3$Loc   & 0.0 & 0.0 & 0.0 & 0.0 & 0.0 & 4.2 \\
      & PALMS      & 0.0 & 0.0 & 11.1 & 11.1 & 0.0 & 0.0 \\
      & PALMS+     & 0.0 & 8.3 & 8.3 & 8.3 & 8.3 & 8.3 \\
      & PALMS+$^*$ & 4.2 & 8.3 & 12.5 & 12.5 & 12.5 & 16.7 \\
    \midrule
    \multirow{4}{*}{3}
      & F$^3$Loc   & 0.0 & 0.0 & 0.0 & 0.0 & 0.0 & 0.0 \\
      & PALMS      & 0.0 & 0.0 & 0.0 & 0.0 & 0.0 & 0.0 \\
      & PALMS+     & 0.0 & 11.1 & 33.3 & 33.3 & 44.4 & 50.0 \\
      & PALMS+$^*$ & 0.0 & 33.3 & 50.0 & 50.0 & 50.0 & 55.6 \\
    \midrule
    \multirow{4}{*}{4}
      & F$^3$Loc   & 0.0 & 0.0 & 0.0 & 0.0 & 0.0 & 0.0 \\
      & PALMS      & 5.6 & 5.6 & 5.6 & 5.6 & 5.6 & 11.1 \\
      & PALMS+     & 0.0 & 16.7 & 33.3 & 33.3 & 33.3 & 33.3 \\
      & PALMS+$^*$ & 5.6 & 22.2 & 38.9 & 38.9 & 38.9 & 38.9 \\
    \bottomrule
  \end{tabular}}
  \caption{Localization accuracy on the custom dataset under the full-view observation setting. This table shows per-building metrics for all methods. PALMS+$^*$ is PALMS+ with masked depths.}
\end{table}

\begin{table}[h]
  \centering
  \small
  \resizebox{0.6\textwidth}{!}{
  \renewcommand{\arraystretch}{1.2}
  \begin{tabular}{l|l|cccccc}
    \toprule
     \multirow{2}{*}{\textbf{Building}} & \multirow{2}{*}{\textbf{Method}} & \multicolumn{6}{c}{Localization Accuracy (\%)} \\
     \cmidrule(lr){3-8}
     & & 0.1m & 0.5m & 1m & 1m $30^\circ$ & 2m & 5m \\
    \midrule
    \multirow{4}{*}{1}
      & F$^3$Loc   & 0.0 & 0.0 & 0.0 & 0.0 & 0.0 & 0.0 \\
      & PALMS      & 5.3 & 10.5 & 10.5 & 10.5 & 10.5 & 15.8 \\
      & PALMS+     & 0.0 & 26.3 & 31.6 & 31.6 & 31.6 & 31.6 \\
      & PALMS+$^*$ & 0.0 & 26.3 & 36.8 & 36.8 & 42.1 & 47.4 \\
    \midrule
    \multirow{4}{*}{2}
      & F$^3$Loc   & 0.0 & 0.0 & 0.0 & 0.0 & 0.0 & 0.0 \\
      & PALMS      & 0.0 & 0.0 & 0.0 & 0.0 & 0.0 & 0.0 \\
      & PALMS+     & 0.0 & 4.2 & 12.5 & 12.5 & 12.5 & 12.5 \\
      & PALMS+$^*$ & 0.0 & 8.3 & 12.5 & 12.5 & 12.5 & 12.5 \\
    \midrule
    \multirow{4}{*}{3}
      & F$^3$Loc   & 0.0 & 0.0 & 0.0 & 0.0 & 0.0 & 0.0 \\
      & PALMS      & 0.0 & 0.0 & 0.0 & 0.0 & 0.0 & 0.0 \\
      & PALMS+     & 0.0 & 5.6 & 11.1 & 11.1 & 16.7 & 16.7 \\
      & PALMS+$^*$ & 0.0 & 0.0 & 11.1 & 11.1 & 11.1 & 11.1 \\
    \midrule
    \multirow{4}{*}{4}
      & F$^3$Loc   & 0.0 & 0.0 & 0.0 & 0.0 & 0.0 & 0.0 \\
      & PALMS      & 5.6 & 11.1 & 11.1 & 11.1 & 11.1 & 11.1 \\
      & PALMS+     & 5.6 & 11.1 & 22.2 & 22.2 & 22.2 & 22.2 \\
      & PALMS+$^*$ & 0.0 & 16.7 & 27.8 & 27.8 & 33.3 & 33.3 \\
    \bottomrule
  \end{tabular}}
  \caption{Localization accuracy on the custom dataset under the partial-view observation setting. This table shows per-building metrics for all methods. PALMS+$^*$ is PALMS+ with masked depths.}
\end{table}

\begin{table}[h]
  \centering
  \small
  \resizebox{0.6\textwidth}{!}{
  \renewcommand{\arraystretch}{1.2}
  \begin{tabular}{l|l|cccccc}
    \toprule
     \multirow{2}{*}{\textbf{Building}} & \multirow{2}{*}{\textbf{Method}} & \multicolumn{6}{c}{Localization Accuracy (\%)} \\
     \cmidrule(lr){3-8}
     & & 0.1m & 0.5m & 1m & 1m $30^\circ$ & 2m & 5m \\
    \midrule
    \multirow{4}{*}{1}
      & F$^3$Loc   & 0.0 & 0.0 & 0.0 & 0.0 & 0.0 & 5.3 \\
      & PALMS      & 0.0 & 5.3 & 5.3 & 5.3 & 5.3 & 15.8 \\
      & PALMS+     & 5.3 & 5.3 & 15.8 & 15.8 & 15.8 & 15.8 \\
      & PALMS+$^*$ & 0.0 & 5.3 & 21.1 & 21.1 & 21.1 & 26.3 \\
    \midrule
    \multirow{4}{*}{2}
      & F$^3$Loc   & 0.0 & 0.0 & 0.0 & 0.0 & 0.0 & 4.2 \\
      & PALMS      & 0.0 & 0.0 & 0.0 & 0.0 & 0.0 & 0.0 \\
      & PALMS+     & 0.0 & 4.2 & 4.2 & 4.2 & 4.2 & 4.2 \\
      & PALMS+$^*$ & 0.0 & 8.3 & 8.3 & 8.3 & 8.3 & 8.3 \\
    \midrule
    \multirow{4}{*}{3}
      & F$^3$Loc   & 0.0 & 0.0 & 0.0 & 0.0 & 0.0 & 0.0 \\
      & PALMS      & 0.0 & 0.0 & 0.0 & 0.0 & 0.0 & 0.0 \\
      & PALMS+     & 0.0 & 5.6 & 5.6 & 5.6 & 5.6 & 11.1 \\
      & PALMS+$^*$ & 0.0 & 5.6 & 5.6 & 5.6 & 5.6 & 11.1 \\
    \midrule
    \multirow{4}{*}{4}
      & F$^3$Loc   & 0.0 & 0.0 & 0.0 & 0.0 & 5.6 & 5.6 \\
      & PALMS      & 0.0 & 0.0 & 0.0 & 0.0 & 0.0 & 0.0 \\
      & PALMS+     & 0.0 & 11.1 & 22.2 & 22.2 & 27.8 & 27.8 \\
      & PALMS+$^*$ & 0.0 & 16.7 & 22.2 & 22.2 & 22.2 & 27.8 \\
    \bottomrule
  \end{tabular}}
  \caption{Localization accuracy on the custom dataset under the single-view observation setting. This table shows per-building metrics. PALMS+$^*$ is PALMS+ with masked depths.}
\end{table}

\section{Additional Qualitative Examples}

\begin{figure}[h]
    \centering
    \resizebox{0.9\textwidth}{!}{\includegraphics[width=\linewidth]{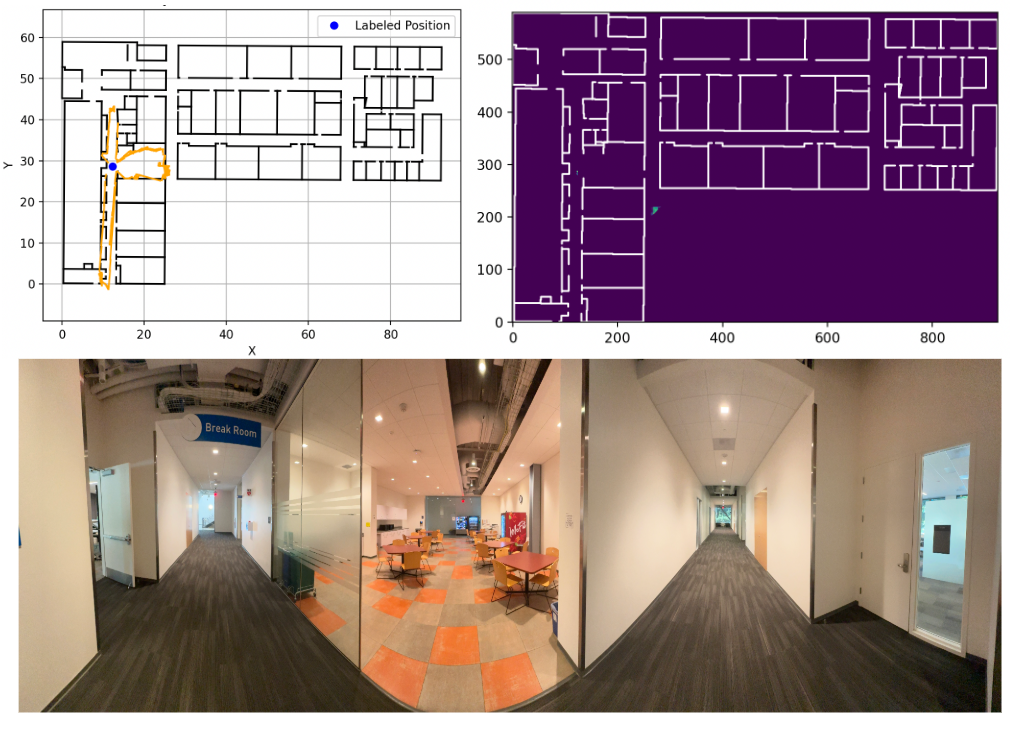}}
    \caption{Distinctive structure is well captured by the system, resulting in relatively optimal performance where only a couple of peaks are present. A strong, concentrated peak at the ground-truth location can be seen in this example.}
    \label{fig:SVC}
\end{figure}

\begin{figure}[h]
    \centering
    \resizebox{0.9\textwidth}{!}{\includegraphics[width=\linewidth]{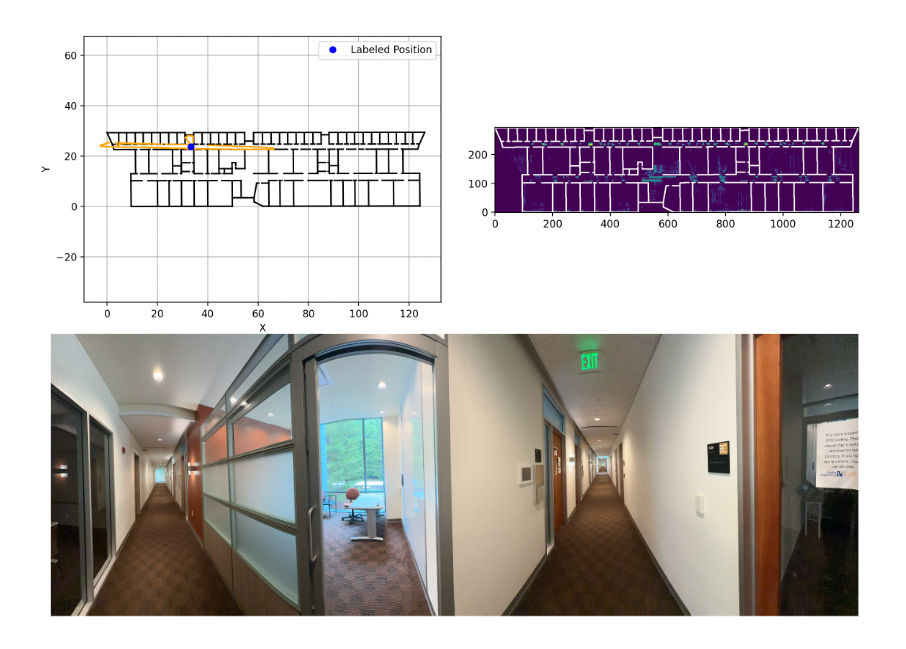}}
    \caption{Long hallways are common, which results in less optimal performance compared to \autoref{fig:SVC}. Multiple peaks are present, reflecting higher geometric ambiguity.}
\end{figure}

\begin{figure}[h]
    \centering
    \resizebox{0.9\textwidth}{!}{\includegraphics[width=\linewidth]{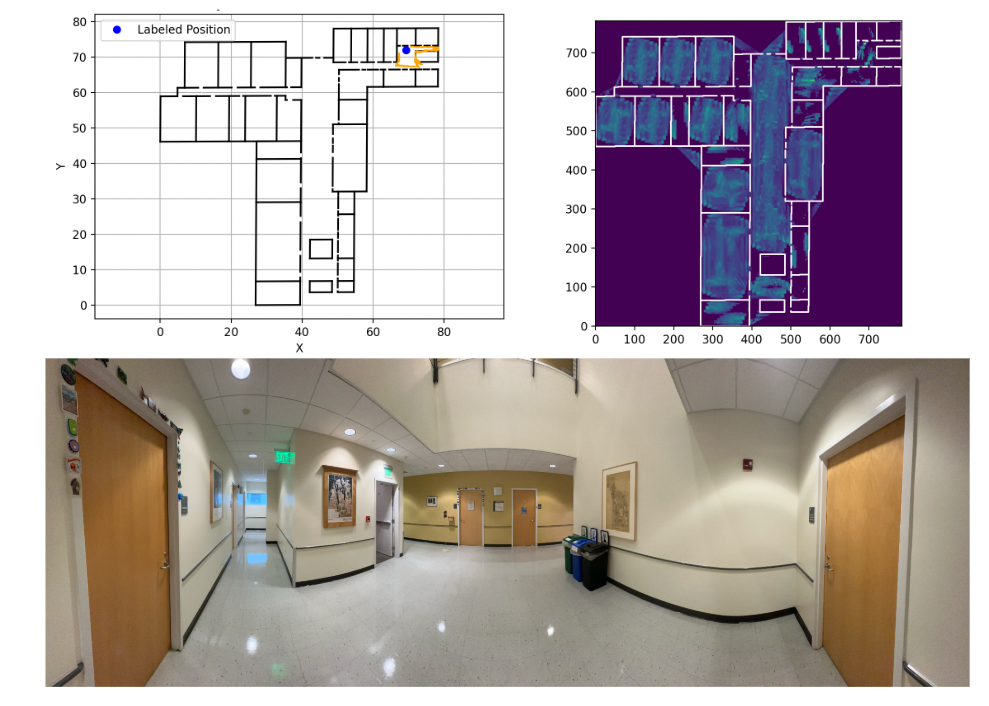}}
    \caption{Observation exhibits a simple box-like structure, which results in low performance and higher ambiguity, as shown by the heatmap.}
\end{figure}

\section{Dataset Specs}

\subsection{Privacy}
To protect privacy, all images are preprocessed to remove sensitive content. Individuals appearing in photographs were manually blurred out to prevent inadvertent identification, and computer screens were similarly blurred to avoid inadvertent exposure of private or confidential information.

\subsection{Distribution of Observation Points}
\begin{figure}[t]
    \begin{subfigure}[b]{0.45\textwidth}
        \centering
        \includegraphics[width=\linewidth]{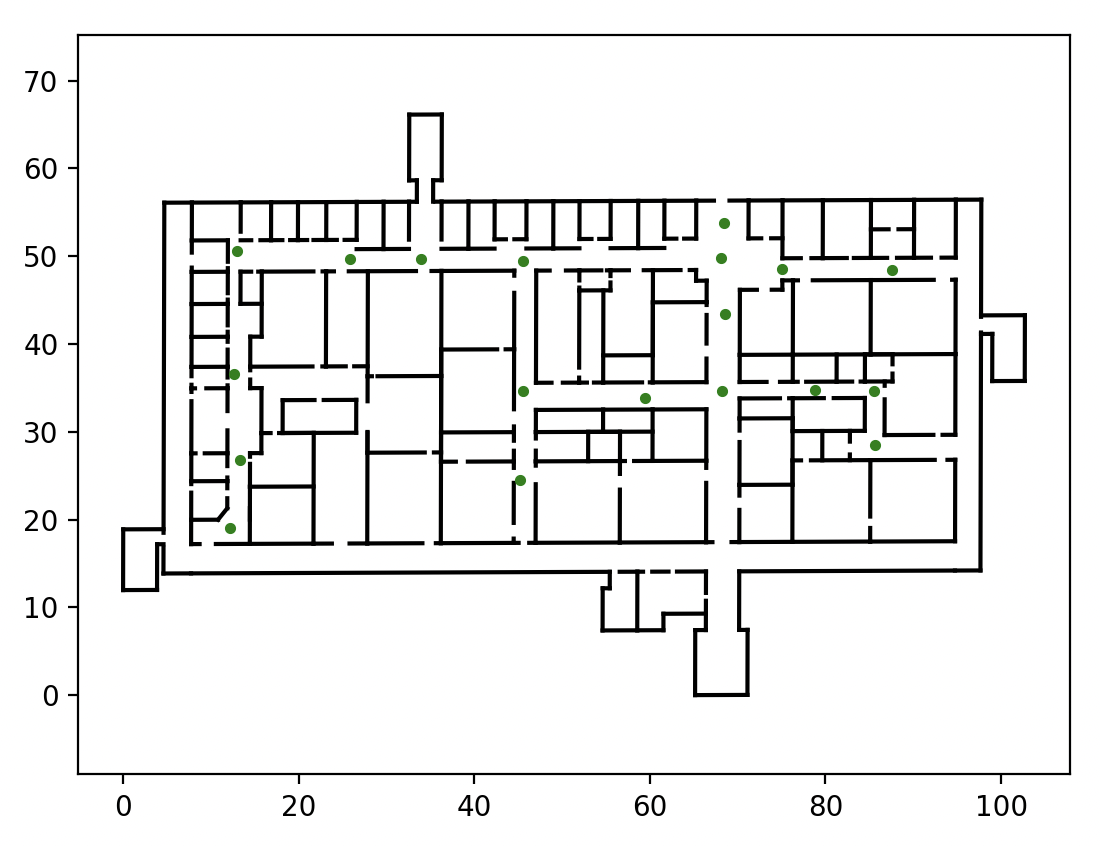}
        \caption{Building 1}
    \end{subfigure}
    \hfill
    \begin{subfigure}[b]{0.45\textwidth}
        \centering
        \includegraphics[width=\linewidth]{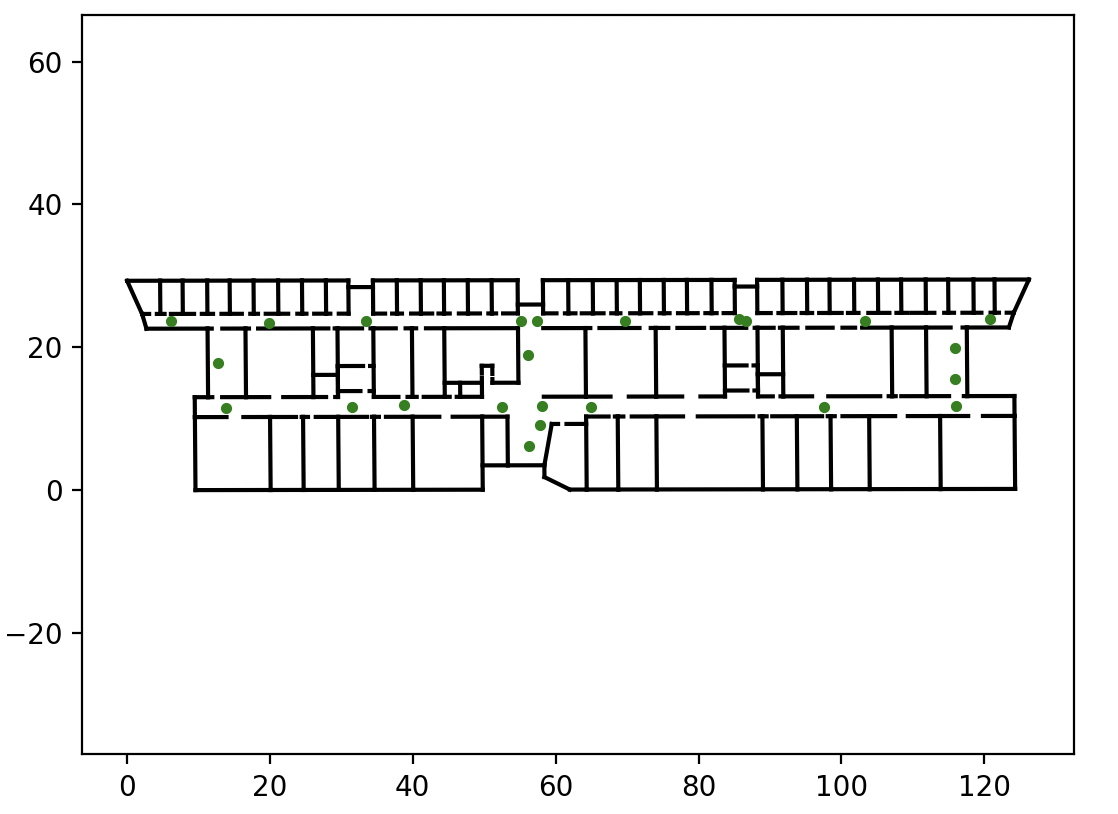}
        \caption{Building 2}
    \end{subfigure}

    \vspace{0.5cm}

    \begin{subfigure}[b]{0.45\textwidth}
        \centering
        \includegraphics[width=\linewidth]{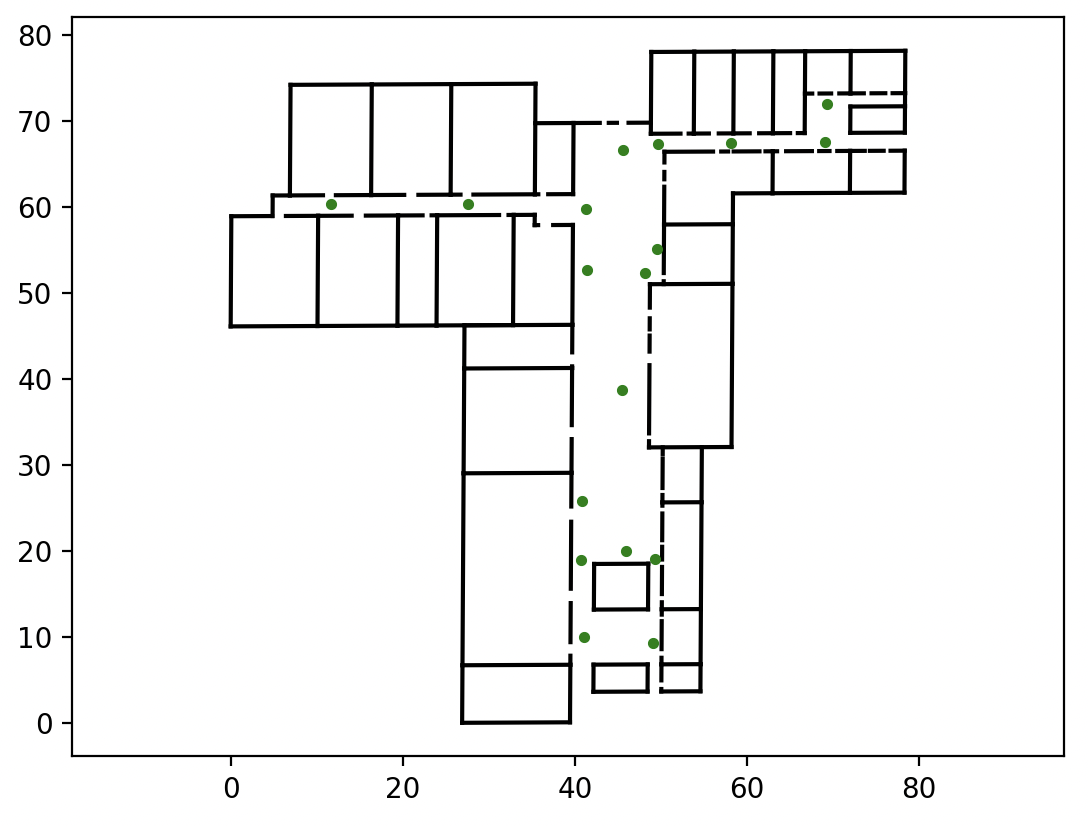}
        \caption{Building 3}
    \end{subfigure}
    \hfill
    \begin{subfigure}[b]{0.45\textwidth}
        \centering
        \includegraphics[width=\linewidth]{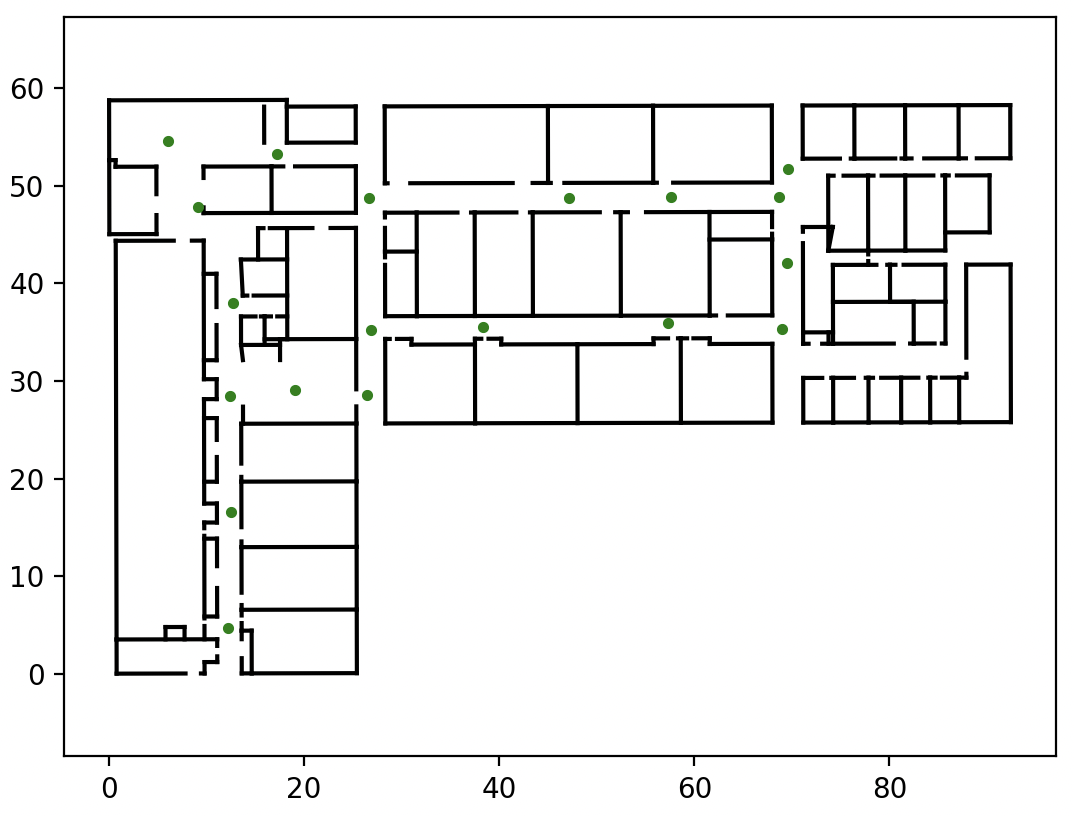}
        \caption{Building 4}
    \end{subfigure}

    \caption{The figures above show all the points at which data were recorded in each of the buildings. We intentionally included a mixture of locations with different levels of ambiguity. Building 2 is the most difficult to localize among all the buildings because it consists mostly of long, homogeneous hallways.}
    \label{fig:four_images}
\end{figure}

See \autoref{fig:four_images}.

\end{document}